\documentclass[10pt,twocolumn,letterpaper]{article}

\usepackage[pagenumbers]{cvpr} 

\usepackage[dvipsnames]{xcolor}

\definecolor{cvprblue}{rgb}{0.21,0.49,0.74}
\usepackage[pagebackref,breaklinks,colorlinks,citecolor=cvprblue]{hyperref}
\usepackage{color}
\usepackage{hyperref}
\usepackage{url}
\usepackage{graphicx}
\usepackage{booktabs}
\usepackage{multirow} 
\usepackage{caption}
\usepackage{lipsum}
\usepackage{tabularx}
\usepackage{float}
\usepackage{stfloats}
\usepackage{makecell}

\definecolor{darkred}{rgb}{0.7, 0.0, 0.0}
\definecolor{darkgreen}{rgb}{0.0, 0.42, 0.24}
\definecolor{darkblue}{rgb}{0.10, 0.17, 0.8}
\definecolor{LQY_color}{RGB}{66, 84, 245}

\def\paperID{7858} 
\def\confName{CVPR}
\def\confYear{2024}

\title{GenNBV: Generalizable Next-Best-View Policy for Active 3D Reconstruction}

\author{Xiao Chen$^{1,2}$\quad Quanyi Li$^{1}$\quad Tai Wang$^{1}$ \quad Tianfan Xue$^{2,*}$ \quad Jiangmiao Pang$^{1,*}$ \vspace{0.1cm} \\
$^1$Shanghai AI Laboratory \quad $^2$The Chinese University of Hong Kong \\
{\tt\small \{cx123,tfxue\}@ie.cuhk.edu.hk \quad quanyili0057@gmail.com \quad \{wangtai,pangjiangmiao\}@pjlab.org.cn} \vspace{0.2cm} \\
\textbf{Project page}: \href{https://gennbv.github.io/}{gennbv.github.io}
}

\begin{document}





\twocolumn[{%
    \maketitle
    \begin{figure}[H]
        \hsize=\textwidth
        \centering
        \vspace{-40pt}
        \includegraphics[width=1.0\textwidth]{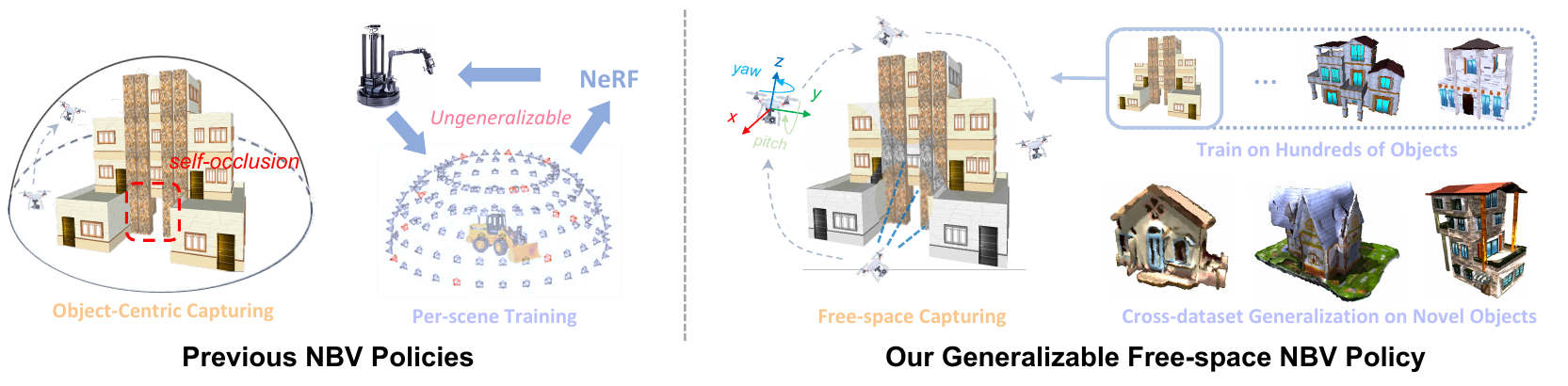}
        \caption{To determine the best view for 3D reconstruction, previous methods only chose from hand-crafted action space or based on object-centric capturing, lacking the ability to generalize to unforeseen scenes (Left). With our end-to-end trained, generalized free-space policy, it can generalize to unseen objects, enabling the captured drone to image from any viewpoint (Right).}
        \label{fig:teaser}        
    \end{figure}
}]

\vspace{-2cm}
\renewcommand{\thefootnote}{\fnsymbol{footnote}}
\footnotetext[1]{Corresponding author.}



\begin{abstract}
    \vspace{-0.2cm}
    While recent advances in neural radiance field enable realistic digitization for large-scale scenes, the image-capturing process is still time-consuming and labor-intensive.
    Previous works attempt to automate this process using the Next-Best-View (NBV) policy for active 3D reconstruction.
    However, the existing NBV policies heavily rely on hand-crafted criteria, limited action space, or per-scene optimized representations. These constraints limit their cross-dataset generalizability.
    To overcome them, we propose \textbf{GenNBV}, an end-to-end generalizable NBV policy. Our policy adopts a reinforcement learning (RL)-based framework and extends typical limited action space to 5D free space. It empowers our agent drone to scan from any viewpoint, and even interact with unseen geometries during training.
    To boost the cross-dataset generalizability, we also propose a novel multi-source state embedding, including geometric, semantic, and action representations.
    We establish a benchmark using the Isaac Gym simulator with the Houses3K and OmniObject3D datasets to evaluate this NBV policy.
    Experiments demonstrate that our policy achieves a 98.26\% and 97.12\% coverage ratio on unseen building-scale objects from these datasets, respectively, outperforming prior solutions.

\end{abstract}
\section{Introduction}
\label{sec:introduction}
\vspace{-0.1cm}
Recent advances in 3D reconstruction~\citep{park2019deepsdf, sun2021neuralrecon, sun2022neuconw} and neural rendering~\citep{mildenhall2020nerf, sun2022direct, zhang2022nerfusion} have significantly enhanced the quality of 3D digitization of large scenes, such as buildings and city landmarks~\citep{hardouin2020next, zhang2021continuous, liu2022learning, xiangli2022bungeenerf, li2023matrixcity}.
However, image-capturing process still remains time-consuming and labor-intensive. To scan a building-scale scene using a drone, it may take several days' effort of a professional team to ensure full coverage. Moreover, even professional pilots may miss some areas at the first scan, leading to multiple rounds for rescanning.

To alleviate the effort in manual scanning, active 3D reconstruction algorithms, especially the Next-Best-View (NBV) policy, have emerged as a promising approach to automate view planning. 
However, typical rule-based NBV policies~\citep{lee2022uncertainty, pan2022activenerf, zhan2022activermap, peralta2020next, guedonscone, ran2023neurar} heavily rely on empirically designed view-selection criteria and hand-crafted action space, such as a hemisphere as shown in Fig.~\ref{fig:teaser}, which limit their generalizability to unseen scenes.
To enhance generalization capacity, pioneering work on learning-based NBV policy has been proposed~\citep{peralta2020next, zhan2022activermap, chenlearning, chaplotlearning}. Nevertheless, they are constrained into impractical setups in real-world scenarios, such as object-centric capturing and limited action space for ground robots, resulting in incomplete 3D reconstruction because of significant self-occlusion.

Therefore, in this work, we study this problem: \textit{``How to design an NBV policy that supports exploration in any space and also can generalize to unseen building-scale geometries?"} 
Building upon the design of previous reinforcement learning-based (RL-based) NBV policies~\citep{peralta2020next, chenlearning, chaplotlearning}, our setup faces the following new challenges that were not fully studied before:
1) Designing an easy-to-generalize action space, rather than object-specific spaces used in previous work, such as hand-crafted candidates or a constrained hemisphere;
2) With a large action space, an informative representation is necessary to guide the policy to efficiently find an optimal capturing trajectory;
3) Proposing generic reward functions and developing the distributed training procedure for better generalizability.

To overcome these two challenges, we propose \textit{GenNBV}, an end-to-end generalizable NBV policy. It has a novel design of action spaces and state embeddings, which allows free-space exploration and ensures a large coverage when applied to unseen objects.
We first extend the former limited action space, like a hemisphere, to 5D free space. As shown in Fig.~\ref{fig:teaser}, this new space is composed of a position subspace of approximately $20m$ x $20m$ x $10m$ and an omnidirectional heading subspace. This free space enables our agent drone to scan from any viewpoint.
Moreover, with a much larger action space, our RL-based policy can be easily generalized to building-scale geometries without concerning the potential scale shift from training to evaluation. In contrast, many hand-crafted designs~\citep{pan2022activenerf, lee2022uncertainty} limit the available viewpoints to a small predefined set, and thus cannot generalize if there is a drastic change in scales.

Second, in order to generalize a well-trained NBV policy to unseen scenes during evaluation, we propose a novel state representation directly from raw sensory inputs to guide the policy. In contrast to the widely used neural radiance field (NeRF)~\citep{pan2022activenerf, zhan2022activermap, lee2022uncertainty, smith2022uncertainty,ran2023neurar}, our representation does not need time-consuming per-scene optimization. 
Specifically, our NBV policy incorporates a multi-source state embedding, which comprises three key components: a novel geometric embedding, a semantic embedding, and an action embedding. The geometric embedding is derived from multi-view depth maps and represents an encoded probabilistic 3D grid, while the semantic embedding is extracted from RGB images, and the action embedding is an encoded viewpoint sequence.
In contrast to 2D representations used in previous RL-based NBV policies~\citep{chenlearning, chaplotlearning}, our multi-source state embedding, extracted from a probabilistic 3D grid, is a more comprehensive representation that supports a robust prediction of next viewpoints.

To validate the effectiveness of GenNBV, we construct a benchmark for training and evaluation with Houses3K~\citep{houses3k} dataset from the NVIDIA Isaac Gym~\citep{makoviychuk2021isaac} simulator. We further test its generalization ability on novel buildings and object categories from OmniObject3D~\citep{wu2023omniobject3d} and Objaverse~\citep{deitke2023objaverse} datasets.
Our evaluation metrics quantify the completeness, accuracy, and efficiency of active 3D reconstruction.
To consolidate completeness and efficiency into a single number, we also propose a metric, the Area Under the Curve (AUC) of the coverage. 
Compared to heuristic~\citep{isler2016information}, information gain-based~\citep{isler2016information, lee2022uncertainty, zhan2022activermap}, and RL-based baselines~\citep{peralta2020next}, our method achieves the best result under different metrics.
We also provide visualizations to demonstrate the generalizability of GenNBV.
A demo video is attached in the appendix to further illustrate our method.

\section{Related Work}
\label{sec:related work}
    \noindent\textbf{Traditional 3D Reconstruction.}
    Photometry and geometry are two crucial aspects of reconstruction evaluation. Neural implicit representations~\citep{park2019deepsdf, mildenhall2020nerf} have shown progress in photometric rendering but faces challenges like time-consuming optimization and poor generalization, hindering its application in real-time reconstruction scenarios like 3D SLAM~\citep{sucar2021imap, iSDF2022, zhu2022nice, rosinol2022nerf}. Geometry, in contrast, typically corresponds to 3D representations such as point clouds and 3D mesh and is directly related to issues like collision avoidance. Considering these properties, researchers tend to focus on geometric reconstruction in practical applications and we also follow this stream.

    \noindent\textbf{Active 3D Reconstruction.}
    Active 3D reconstruction is a promising field that has not been thoroughly benchmarked yet. The pipeline of active 3D reconstruction frameworks alternates between inferring optimal viewpoints, capturing new data, and updating the rebuilt 3D model. With the optimal sequence of key viewpoints, a continuous planning path can be generated using the classic planners, such as Fast Marching Method (FMM)~\citep{sethian1996fast}.  

    Existing works can be pivotally differentiated based on the paradigm of NBV policies: rule-based or learning-based.
    Rule-based approaches like uncertainty-driven works~\citep{isler2016information, lee2022uncertainty, zhan2022activermap, ran2023neurar} typically yield the next best view from hand-crafted rules from scene representations, which tends to overfit specific scenes. 
    Most learning-based policies~\citep{chenlearning, chaplotlearning, peralta2020next} use a deep reinforcement learning algorithm like PPO~\citep{schulman2017proximal} to sequentially predict the optimal viewpoints based on observation. They must obtain feedback from task-related rewards such as coverage ratio~\citep{peralta2020next, guedonscone} and optimize during massive iteration with task environment.

    The action space is designed based on the paradigm of NBV policies. Most rule-based NBV policies select views from a limited action space, such as a set composed of only one hundred candidate viewpoints~\citep{pan2022activenerf}, a tiny pre-defined space like a hemisphere~\citep{lee2022uncertainty, zhan2022activermap, guedonscone, ran2023neurar}, making it possible to overlook some important viewpoints due to unavailability. Even though learning-based categories further explore the larger action space like a 2D plain~\citep{chenlearning, chaplotlearning} or a constrained 3D space~\citep{peralta2020next}, their limited action space still prevents them from capturing sufficient details for 3D reconstruction.
    
    Scene representation built from history observations, which directly provide reconstruction progress to NBV policies, is also a crucial aspect of the active 3D reconstruction framework. Previous works have explored visual representations for NBV policies, such as TSDF~\citep{hardouin2020next}, neural radiance field~\citep{adamkiewicz2022vision}, and 2D BEV maps~\citep{chaplotlearning, ye2022multi, guo2022asynchronous}. However, implicit representations are hard to jointly with learning-based NBV policies, and 2D BEV map lacks sufficient information for large-scale outdoor scenes that contain numerous geometric details in 3D space. 

    In our benchmark, we select Scan-RL~\citep{peralta2020next} as our main RL-based baseline. Compared to it, we introduce a much larger action space and informative visual representations for better generalizability. Additionally, we don't benchmark the mentioned baselines~\citep{chenlearning,chaplotlearning}, mainly due to their reliance on pretraining from human demonstration using imitation learning, while our focus lies on learning through iterations and end-to-end training. Different from our 3D free-space movement, they only explore the traversable 2D areas of indoor scenes using ground robots, thus limiting their motion space in 2D.

    \begin{figure*}[t!]
      \centering
      \includegraphics[width=1\textwidth]{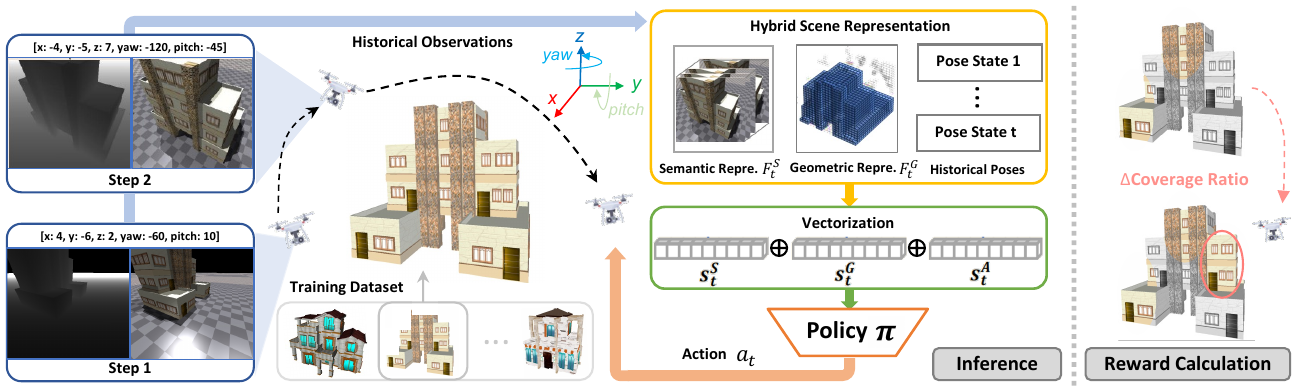}
      \caption{Overview of our proposed framework GenNBV. Our end-to-end policy takes the historical multi-source observations as input,  transforms them into a more informative scene representation, and produces the next viewpoint position. A reward signal will be returned at training time to optimize the end-to-end policy for maximizing the expected cumulative reward in one episode. Specifically, the signal is the increased coverage ratio after collecting a new viewpoint.}
      \label{fig:method}
      \vspace{-10pt}
    \end{figure*}

    \noindent\textbf{Generalizable Reinforcement Learning.}
    Increasing the diversity of training data leads to better generalizability~\citep{cobbe2020leveraging}.
    In the robotics field, domain randomization~\citep{tobin2017domain} allows legged robots to walk on various terrains absent from the training environment. For end-to-end driving policy, training in large-scale synthetic and realistic scenarios improves the safety of autonomous cars~\citep{li2022metadrive, li2023scenarionet} in unseen test scenarios.
    The advanced work~\citep{devrim2017reinforcement} presents an RL-based formulation for the view planning problem. Based on similar frameworks of active 3D reconstruction, Scan-RL~\citep{peralta2020next} and SCONE~\citep{guedonscone} are two pioneering works showing the generalizability of their frameworks, while both of them suffer from the constrain of viewpoint sampling in limited space and lack of in-depth analysis.
    In this work, we tackle the aforementioned issues by predicting NBV in free 3D space with a learning-based planner, which is trained with a dataset containing buildings in various shapes and poses for acquiring generalizability for unseen buildings.

\vspace{1pt}
\section{Methodology}
\label{sec:method}
    In this section, we deliver our active 3D reconstruction framework GenNBV, especially the pivotal Next-Best-View policy. An overview of our framework is illustrated in Fig.~\ref{fig:method}.
    Firstly, we formulate the NBV problem as a Markov Decision Process (MDP), with a novel design of observations (blue boxes in Fig.~\ref{fig:method}) and action space in Sec.~\ref{sec:method_setup}.
    Next, we elaborate our end-to-end NBV policy $\pi$ (orange box in  Fig.~\ref{fig:method}) in Sec.~\ref{sec:method_state}. Inspired by Curl~\citep{laskin2020curl}, we point out that generalizability greatly depends on how to extract embeddings (green box in Fig.~\ref{fig:method}), which reflect the reconstruction progress, from raw sensory observations like RGB images and camera poses.
    In Sec.~\ref{sec:method_reward}, we introduce the reward function (right-hand side of  Fig.~\ref{fig:method}) reflecting the optimization objective and the details of policy optimization.

\vspace{4pt}
\subsection{Formulation of the Next-Best-View Problem}
\label{sec:method_setup}
    We formulate the NBV problem as learning an optimal policy $\pi$ that controls the capturing process, such that enough information is captured for large-scale scene reconstruction, with limited decision-making budgets.
    As capturing, transferring, and processing a large set of captured images introduce significant computational costs, we also want to design a policy that also minimizes the number of captured images. Therefore, our policy only captures sparse keyframes that sufficiently record all details of objects.

    As shown in Fig.~\ref{fig:method}, at each time step $t$, the agent receives a visual observation $o_t$, takes an action $a_t$, infers the action at the next time step $t+1$ to move to a new location, and then receives a new visual observation, repeating this interaction with the environment until the episode ends. Our simulated agent is embodied as CrazyFlie~\citep{giernacki2017crazyflie}, a type of unmanned aerial vehicle equipped with various sensors, including an RGB-D camera and an IMU, to execute data collection for reconstruction. We discussed the details of the observation space and action space below.

    \vspace{1pt}
    \noindent\textbf{Observation Space.} As shown in the left column of Fig.~\ref{fig:method}, at each time step, the agent receives an RGB image $I_t$, a depth map $D_t$, and a state vector including the heading (yaw and pitch) and position (x, y, z) of the onboard camera.
    The observation $o_t$ (yellow box in Fig.~\ref{fig:method}) consists of all previously captured images in one episode, historical actions, and the current captures and actions. With this information as input, the policy network can estimate the progress of data collection and determine where to scan next.

    \vspace{1pt}
    \noindent\textbf{Action Space.} Unlike most previous NBV algorithms, we use larger action space in order to cover all details of objects. Specifically, we use the camera location and camera angle as our action space, which is a 5-dimension vector consisting of 3D position coordinates and 2D rotation angles (yaw-axis and pitch-axis). We restrict the roll-axis rotation as it is not supported in all drone platforms.

\subsection{Generalizable State Embedding}
\label{sec:method_state}
    To ensure that the policy learned on one set of 3D objects can generalize to objects with different appearances and structures, a smart state embedding of raw sensor observations is needed to capture invariant features across different objects. Previous methods ~\citep{peralta2020next, chaplotlearning} only extract representations in 2D space, which are very sensitive to appearance changes, and our experiments have shown that policies only trained on 2D features may not be generalized to different objects. Instead, we propose a multi-source representation that has better generalizability.

    Specifically, we first build two mid-level representations that can better model the relationship between the objects and our agent: a 3D geometric representation $F^G$ from depth maps and a semantic representation $F^S$ from RGB images. Then we encode these representations and concatenate them with a pose embedding into state embedding $s_t$, guiding the NBV policy for subsequent decision-making. The overall encoder for state embedding is shown in Fig.~\ref{fig:method}, and details of each representation are discussed below.

    \vspace{1pt}
    \noindent\textbf{Geometric Representation}\quad  
    One simple way to record the geometry of a 3D object is using a binary 3D occupancy map~\citep{chaplotlearning}, where the value of each cell in the 3D cube indicates whether a cell contains a 3D object or not. However, since the 3D representation is gradually updated with newly captured data, this simple binary occupancy map cannot differentiate a real unoccupied cell from an unscanned cell. An unscanned cell is a strong indicator that the agent should capture more data in this region, while no further scan is needed for a real unoccupied cell.
    
    Previous works~\citep{chaplotlearning, ye2022multi, guo2022asynchronous} simplify the 3D scene representation into a 2D Bird's Eye View map for actively reconstructing indoor scenes. However, the 2D BEV map lacks sufficient information for our large-scale outdoor scenes that contain many geometric details in 3D space. In addition, their wheeled robotic platforms are constrained to freely scan and represent these outdoor scenes.
    Therefore, to model the scanning process in 3D free space, we employ the probabilistic 3D grid~\citep{thrun2002probabilistic} as our geometric 3D representation, which records the probability of each 3D voxel being captured or not. Specifically, we first obtain a 3D point cloud in the world coordinate by back-projecting all 2D pixels to 3D points, using the depth map $D_t$, camera intrinsic parameters, and camera pose $a_t$. By voxelizing the obtained point cloud, we then build a 3D occupancy grid that explicitly indicates the binary state (occupied or free) in this 3D space. 
    Subsequently, we represent this 3D grid as a probabilistic occupancy grid $F^G_t$ and extend the state space of voxels with three states (occupied, free, unknown).

    During each scanning (one episode in RL), at each step $t+1$, we update the probabilistic occupancy grid $F_{t+1}^G$ based on the preceding grid $F_{t}^G$ and the current observation. Specifically, the grid is updated through Bresenham's line algorithm~\citep{bresenham1965algorithm}, which casts the ray path in 3D space between the camera viewpoint and the endpoints among the point cloud back-projected from depth $D_{t+1}$. Following the classical occupancy grid mapping algorithm~\citep{thrun2002probabilistic}, we have the log-odds formulation of occupancy probability:
        \begin{equation}
        \log \mathrm{Odd}(v_i|z_j) = \log \mathrm{Odd}(v_i) + C,
        \end{equation}
    where $v_i$ denotes the occupancy probability of $i^{th}$ voxel in the grid $F^G_t$, $z_j$ is the measurement event that $j^{th}$ camera ray passes through this voxel and C is an empirical constant. The derivation can be found in Appendix.
    Thus, we update the log-odds occupancy probability of each voxel in the grid $F^G_t$ by adding a constant one time when a single camera ray passes through this voxel. Note that the probabilistic occupancy grid $F^G$ is continually updated within one episode. Finally, the occupancy status of voxels is classified into three categories: unknown, occupied, and free, using preset probability thresholds.


    \vspace{1pt}
    \noindent\textbf{Semantic Representation.} 
    Geometric representation enables agents to comprehend spatial occupancy, yet it's insufficient for perceiving the environment. 
    For example, when observing a hole in an object, the agent may struggle to differentiate between incomplete scanning and the actual presence of a hole in the object. In such cases, the semantic information contained in the captured RGB images can help the agent distinguish between these two scenarios.
    
    To provide semantic information, we employ a preprocessing module that takes as input the current frame of RGB image $I_t$ and preceding $K$ frames and converts these frames $[I_t, I_{t-1}, ..., I_{t-k}]$ to grayscale, and concatenate them as output, following \cite{peralta2020next}. Then the preprocessed frames are fed into a two-layer convolutional network for extracting the semantic representation $F^S_t$.

    \vspace{1pt}
    \noindent\textbf{State Embedding.}
    To further combine the semantic and geometric embeddings, we first encode them to $s^S_t=f^S(F^S_t)$ and $s^G_t=f^G(F^G_t)$ where $f^*$ are learnable networks $\mathrm{Linear}(\mathrm{Flatten}(x))$, as shown in Fig.~\ref{fig:method}. Subsequently, we combine them with the historical action embedding $s^A_t=\mathrm{Linear}(a_{1:t})$ to generate the final state embedding $s_t$, as the input to the policy network. This process can be formulated as: $s_t = \mathrm{Linear}(s^G_t; s^S_t; s^A_t)$.

    \vspace{1pt}
    \noindent\textbf{Policy Network.} Taking the state embedding $s_t$ as input, the policy network is a 3-layer multi-layer perceptron network (MLP) whose output is used to parameterize a normal distribution over action space. In this way, the action can be drawn from the stochastic policy $a \sim \pi(\cdot |o_t)$.

\subsection{Reward Function and Optimization}
\label{sec:method_reward}
    We train the end-to-end policy with reinforcement learning (RL) and hence design reward functions to precisely reflect the task objective for 3D reconstruction. The policy is optimized with proximal policy optimization~\cite{schulman2017proximal} (PPO) for parallelizing sampling.

     \vspace{1pt}
     \noindent\textbf{Reward Functions.}
     With the occupancy probability $F^G_t$ at time step $t$, we can threshold each voxel with an empirical bound to determine if it is occupied.
     This discrimination process outputs a binary occupancy grid with $\Tilde{N}_t$ voxels being occupied, which is used to calculate the coverage ratio: 
        \begin{equation}
        \text{CR}_t = \frac{\Tilde{N}_t}{N^*} \cdot 100\% ,
        \end{equation}
    where $N^*$ is the number of ground-truth occupied voxels representing the surface of objects.
    To encourage our NBV policy to cover as many unseen areas of objects as possible, we use the difference of coverage ratio (CR) between two consecutive steps as the main reward function $r^{CR}$:
    \begin{equation}
    r^\text{CR}_{t+1} = \text{CR}_{t+1} - \text{CR}_t.
    \end{equation}

    In free-space exploration, we also need to avoid collision. Previous limited-space agents~\citep{lee2022uncertainty, zhan2022activermap, guedonscone} do not consider collision avoidance since their search space, like hemisphere, is by-design safe. Thus, we add a negative reward for collision and terminate the episode if a collision happens. We also implement a negative reward when the number of captured keyframes is over an empirical threshold to improve the path efficiency. 

    \vspace{1pt}
    \noindent\textbf{Policy Optimization.}
    Once the reward functions has been specified, the policy can be learned through any off-the-shelf RL algorithm. 
    In this work, we use PPO as thousands of workers can be parallelized to improve the sample efficiency. 
    Specifically, given our parameterized policy $\pi_\theta$, the objective of PPO is to maximize the following function:
        \begin{equation}
        L(\theta) = \mathbb{E}_{t}\left[\frac{\pi_\theta(a_t|s_t)}{\pi_{\theta_\text{old}}(a_t|s_t)} A^{\pi_{\theta_\text{old}}}(s_t, a_t)\right],
        \end{equation}
     where $A^{\pi_{\theta_\text{old}}}(s_t, a_t)$ is the advantage function that measures the value of taking action $a_t$ at state $s_t$ under the current policy $\pi_{\theta_\text{old}}$.
    To prevent significant deviation of the new policy from the old policy, PPO incorporates a clipped surrogate objective function:
        \begin{equation}
            \begin{split}
            L^\text{CLIP}(\theta) =  &\mathbb{E}_{t}[\text{min}(\eta_t(\theta) A^{\pi_{\theta_\text{old}}}(s_t, a_t), \\
                                    &\text{clip}(r_t(\theta), 1 - \epsilon, 1 + \epsilon) A^{\pi_{\theta_\text{old}}}(s_t, a_t))] ,
            \end{split}
        \end{equation}
        where $\eta_t(\theta) = \frac{\pi_\theta(a_t|s_t)}{\pi_{\theta_\text{old}}(a_t|s_t)}$ and $\epsilon$ is a hyper-parameter that controls the size of the trust region.

\vspace{3pt}
\section{Experiments}
\label{sec:exp}

    \begin{table*}[!t]
    \small
    \center
    \setlength\tabcolsep{9pt}
    \renewcommand{\arraystretch}{1.6}
    \captionof{table}{Evaluation results of Next-Best-View policies for active 3D reconstruction on \textbf{Houses3K} and the house category from \textbf{OmniObject3D} (cross-dataset generalization). 
    The number of views is set to 30 and 20 for Houses3K and OmniObject3D, respectively.\\
    ``*": the policy is trained with the Houses3K training set and evaluated with holdout Houses3K test set and OmniObject3D house category. 
    ``\dag": the policy heavily relies on optimized per-scene representation (NeRF), and thus is directly trained and evaluated on testing objects.}
    \label{tab:main_table}

    \begin{tabularx}{\textwidth}{ll|ccc|ccc}
    & & \multicolumn{3}{c|}{\textbf{Houses3K Test Set}} & \multicolumn{3}{c}{\textbf{OmniObject3D}}\\ \hline \noalign{\vskip0.4ex}
    & \textbf{NBV Policy} & \shortstack[l]{\textbf{Mean} \\ \textbf{AUC$~\uparrow$}}  & \shortstack[c]{\textbf{Final Coverage} \\ \textbf{Ratio$~\uparrow$}} & \shortstack[c]{\textbf{Accuracy} \\ (cm) $\downarrow$} & \shortstack[l]{\textbf{Mean} \\ \textbf{AUC$~\uparrow$}} & \shortstack[c]{\textbf{Final Coverage } \\ \textbf{Ratio$~\uparrow$}} & \shortstack[c]{\textbf{Accuracy} \\ (cm) $\downarrow$} \\ \hline

    \parbox[c]{1.2mm}{\multirow{3}{*}{\rotatebox[origin=c]{90}{Heuristic~}}}
    & Random & 48.53\% & 58.24\% & 1.38 & 60.13\% & 73.73\% & 0.62\\
    & Random Hemisphere & 71.19\% & 79.72\% & 0.63 & 74.03\% & 84.74\% & 0.48\\
    & Uniform Hemisphere & 82.91\% & 89.71\% & 0.44 & 83.09\% & 92.90\% & 0.41\\ \hline

    \parbox[c]{1.2mm}{\multirow{2}{*}{\rotatebox[origin=c]{90}{InfoGain~}}}
    & \dag Uncertainty-Guided~\citep{lee2022uncertainty} & 83.13\% & 89.31\% & 0.44 & 69.08\% & 92.85\% & 0.42 \\
    & \dag ActiveRMap~\citep{zhan2022activermap} & 84.86\% & 90.77\% & 0.44 & 69.47\% & 93.15\% & 0.42 \\ \hline

    \parbox[c]{1.2mm}{\multirow{4}{*}{\rotatebox[origin=c]{90}{RL-based~}}}
    & *Scan-RL~\citep{peralta2020next} & 84.49\% & 91.61\% & 0.40 & 82.17\% & 92.53\% & 0.34 \\
    & *Scan-RL w/ Our Repre. & 86.48\% & 92.20\% & \textbf{0.37} & 86.14\% & 93.73\% & 0.35\\
    & *Ours w/ Scan-RL Repre. & 87.39\% & 95.63\% & 0.38 & 86.81\% & 93.64\% & 0.34\\
    & \textbf{*GenNBV (Ours)} & \textbf{91.19\%} & \textbf{98.26\%} & \textbf{0.37} & \textbf{88.63\%} & \textbf{97.12\%} & \textbf{0.33} \\ \hline
    \end{tabularx}
    \vspace{-5pt}
    \end{table*}

    In this section, we conduct experiments on Houses3K~\cite{houses3k}, OmniObject3D~\cite{wu2023omniobject3d}, and Objaverse~\cite{deitke2023objaverse}.
    For our policy and other learning-based policies, we train them on Houses3K training set.
    Then we evaluate all policies on the Houses3K test dataset and the OmniObject3D dataset for quantifying the in-distribution and out-of-distribution generalizability.
    Finally, we show the visualization results on three datasets, demonstrating the effectiveness of the proposed method. 

\subsection{Experimental Setup}
    \noindent\textbf{Simulation Environment.}
    We conduct all experiments in NVIDIA Isaac Gym~\citep{makoviychuk2021isaac}, a physics simulation platform designed for reinforcement learning and robotics research. 
    With GPU-accelerated tensor API and sensor interaction API, we can easily implement customized CUDA kernels to calculate our geometric representation efficiently.
    Moreover, the sensor simulation is efficient. 
    It can run up to 1000 FPS, significantly reducing the training time.
    We create the agent drone CrazyFlie~\citep{giernacki2017crazyflie} and equip it with sensors including RGB-D cameras and IMU. Considering memory limitations, we downsample the image resolution to $400\times 400$. The vertical field of view of this onboard camera is 90\textdegree. We also set up a point light source with a fixed position and constant intensity in Isaac Gym.

    \vspace{1pt}
    \noindent\textbf{Dataset.}
    We conduct our experiments on Houses3K~\citep{houses3k}, OmniObject3D~\citep{wu2023omniobject3d}, Objaverse 1.0~\citep{deitke2023objaverse} and Replica~\citep{straub2019replica}. Our model is trained on large-scale 3D objects from Houses3K. To validate the effectiveness and generalizability of our model, we evaluate it on \textit{unseen} 3D objects from Houses3K and further on batches of \textit{unseen cross-domain} 3D objects from OmniObject3D, Objaverse, and Replica, which include both house and non-house categories.

    \begin{table}[!htbp]
        \center
        \large
        \caption{The cross-dataset generalization for \textbf{non-house categories}. We train the baseline Scan-RL and our GenNBV on Houses3K and generalize them to non-house categories from OmniObject3D and an indoor scene from Replica.}
        \renewcommand{\arraystretch}{1.5}
        \resizebox{1.0\linewidth}{!}{
            \begin{tabular}{l|c|c|ccc}
            \hline
            \label{tab:non_house}
            \vspace{-0.1cm}
            \textbf{Category} & \textbf{Dataset} & \textbf{NBV Policy} & \textbf{AUC $\uparrow$} & \textbf{FCR $\uparrow$} & \textbf{Acc. $\downarrow$} \\ \hline
             \multirow{2}{*}{Animals} & \multirow{2}{*}{OmniObject3D} & Scan-RL & 76.43\% & 84.98\% & 0.17 \\
             & & GenNBV & \textbf{84.28\%} & \textbf{94.07\%} & \textbf{0.16} \\ \hline
             \multirow{2}{*}{Trucks} & \multirow{2}{*}{OmniObject3D}  & Scan-RL & 69.58\% & 76.37\% & 0.06 \\
             & & GenNBV & \textbf{75.54\%} & \textbf{83.47\%} & \textbf{0.03} \\ \hline
             \multirow{2}{*}{Dinosaurs} & \multirow{2}{*}{OmniObject3D}  & Scan-RL & 78.47\% & 86.89\% & 0.88 \\
             & & GenNBV & \textbf{84.24\%} & \textbf{94.03\%} & \textbf{0.81} \\ \hline
             \multirow{2}{*}{Room 0} & \multirow{2}{*}{Replica}  & Scan-RL & 69.21\% & 80.75\% & \textbf{3.12} \\
             & & GenNBV & \textbf{88.53\%} & \textbf{99.16\%} & 3.59 \\ \hline
            \end{tabular}
        }
    \vspace{-15pt}
    \end{table}

    Houses3K contains 3,000 textured 3D building models. They are divided into 12 batches, with each batch featuring 50 distinct geometries and 5 varying textures. We further eliminate poorly designed geometries based on the following two criteria: 1) objects with complex internal structures (as we are mostly focusing on the surface reconstruction), and 2) objects with redundant bases at the bottom. Following these criteria, the training set consists of 256 objects from 6 selected batches, and the test set consists of 50 objects from another batch. All these selected training and test objects have distinct geometries.

    OmniObject3D offers a collection of 6K high-fidelity objects scanned from real-world sources across 190 typical categories. The house category from OmniObject3D for evaluation has 43 diverse objects. In contrast, Objaverse 1.0 comprises a vast library of 818K 3D synthetic objects spanning 21K categories. The diverse and high-quality nature of these datasets makes them ideal for evaluating our models and visualizing the results.

    \begin{figure*}[htbp]
      \centering
      \includegraphics[width=1.0\textwidth]{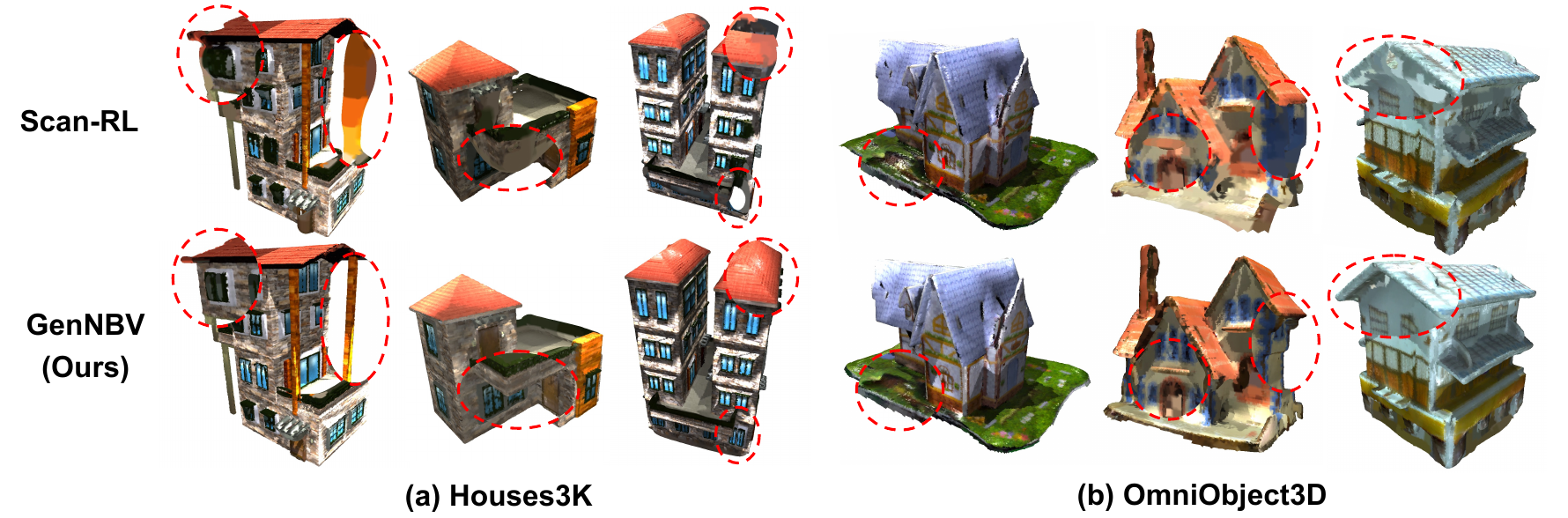}
      \caption{\small The visualization results of unseen 3D objects reconstructed by Scan-RL~\citep{peralta2020next} and our model to compare the generalizability. (a) Unseen buildings from the test set of Houses3K. (b) Unseen buildings from OmniObject3D.
      It's quite obvious that some parts of the model reconstructed by Scan-RL are wrong or missing. For example, the second object in the first row has a pillar in the wrong shape. Scan-RL fails to reconstruct the roof edge for the fourth object from OmniObject3D, as shown in the third row.}
      \label{fig:visual}
      \vspace{-9pt}
    \end{figure*}

    \vspace{1pt}
    \noindent\textbf{Evaluation Metrics.}
    The objective of NBV policies is to capture the most useful information for reconstruction, with the least number of views. We report the \textit{(1) Final Coverage Ratio (\%)}, \textit{(2) Mean AUC (\%)} of coverage ratio and \textit{(3) Reconstruction Accuracy (cm)} along with the consistent number of views in all tables.
    Specifically, most prior works ~\citep{peralta2020next, guedonscone} separately use the coverage ratio (\%) and the number of views to evaluate the reconstruction completeness and efficiency for NBV policies. 
    However, the coverage ratio is highly correlated with the number of views. Thus we propose to unify the number of views to a fixed value for all NBV policies during evaluation and use the area under the curve (AUC) of coverage ratio as the main metric for comparison. 
    Also, we calculate the Chamfer Distance between scanned and ground-truth point clouds to represent the reconstruction accuracy (cm).

    \vspace{1pt}
    \noindent\textbf{Implementation Details.}
    We conduct all experiments in Isaac Gym simulation engine with one NVIDIA Tesla V100 GPU. Our NBV policy is optimized through over 32 million iterations and uses approximately 24 hours of training time on an NVIDIA V100 GPU. All networks are randomly initialized and trained end-to-end. Our implementation refers to the codebase of Legged Gym~\citep{rudinlearning} and the PPO implementation in Stable Baseline3~\citep{raffin2021stable}, which is implemented by PyTorch~\citep{paszke2019pytorch}. The ground-truth point clouds on objects' surfaces are sampled by the Poisson Disk sampling method~\citep{yuksel2015sample} using Open3D API~\citep{Zhou2018}. The Chamfer Distance between the scanned point cloud and the ground-truth point cloud is calculated using PyTorch3D API~\citep{ravi2020pytorch3d}. Please refer to the Appendix for further details and experiments.

\subsection{Performance Comparison}
    \label{sec:main_result}
    \vspace{-2pt}
    To comprehensively demonstrate the effectiveness and generalizability of GenNBV, we design three levels of evaluation experiments: 1) As shown in Table~\ref{tab:main_table}, we show the performance of our NBV policy on the Houses3K test set; 2) We generalize the policy trained on Houses3K to the house category from OmniObject3D that has completely different geometric structures and textures compared to the training set. The quantitative result is at Table~\ref{tab:main_table} and the visualization result is as shown in Fig.~\ref{fig:visual}; 3) We also generalize GenNBV to non-house categories from OmniObject3D and scenes with enormous details from Objaverse and Replica~\citep{straub2019replica} to demonstrate its generalizability potential.

    We implement the following policies as our baselines. 1) \textbf{Random}: This policy randomly generates 5-dim vector $(x, y, z, pitch, yaw)$ among the action space as the next viewpoint. 2) \textbf{Random Hemisphere}: This policy randomly generates the next positions on a pre-defined hemisphere that sufficiently covers all objects of the test set. The headings are constrained to point to the center of the hemisphere. 3) \textbf{Uniform Hemisphere}: All positions are evenly distributed on the previously mentioned hemispheres. 4) \textbf{Uncertainty-Guided Policy}~\citep{lee2022uncertainty}: This NBV policy iteratively selects the next view from a pre-defined viewpoint set according to the uncertainty based on a continually optimized neural radiance field. 5) \textbf{ActiveRMap}~\citep{zhan2022activermap}: This policy also adopts an iterative selection framework, with multiple objectives including information gain, to select the next viewpoints from the candidate set. 6) \textbf{Scan-RL}~\citep{peralta2020next}: This RL-based policy predicts the next viewpoint from a hemisphere space, relying on the historical RGB images. 7) \textbf{Ours with Scan-RL's Representation}: This free-space NBV policy replaces our representation with Scan-RL's representation only extracted from RGB images.

    \begin{table}[ht]
        \center
        \setlength{\tabcolsep}{7.5pt}
        \renewcommand{\arraystretch}{1.5}
        \caption{Ablation studies of representation categories in our framework on unseen Houses3K test set.}
        \label{tab:ablation_repre}
        \resizebox{1.0\linewidth}{!}{
        \begin{tabular}{lccc|cc}
        \toprule
            & \multicolumn{3}{c|}{\textbf{Representation Category}} & \multicolumn{2}{c}{\textbf{Evaluation Metrics}} \\
            \cmidrule(lr){2-4} \cmidrule(lr){5-6}
            & Probabilistic & 5-DoF & Semantic & Mean & Final Coverage \\
            & 3D Grid & Pose & 2D Map & AUC & Ratio \\
            \midrule
            \parbox[c]{1.2mm}{\multirow{3}{*}{\rotatebox[origin=c]{90}{Uni-source~}}}
            & \checkmark & & & 81.06\% & 84.56\% \\ 
            & & \checkmark & & 69.53\% & 76.61\% \\ 
            & & & \checkmark & 81.24\% & 87.90\% \\ 
            \midrule
            \parbox[c]{1.2mm}{\multirow{4}{*}{\rotatebox[origin=c]{90}{Multi-source~}}}
            & \checkmark & \checkmark & & 88.66\% & 96.67\% \\  
            & \checkmark & & \checkmark & 89.77\% & 95.31\% \\  
            & & \checkmark & \checkmark & 88.30\% & 96.29\% \\  
            & \checkmark & \checkmark & \checkmark & \textbf{91.19\%} & \textbf{98.26\%} \\
        \bottomrule
        \end{tabular}}
    \vspace{-20pt}
    \end{table}

    \begin{figure*}[t]
      \centering
      \includegraphics[width=1.0\textwidth]{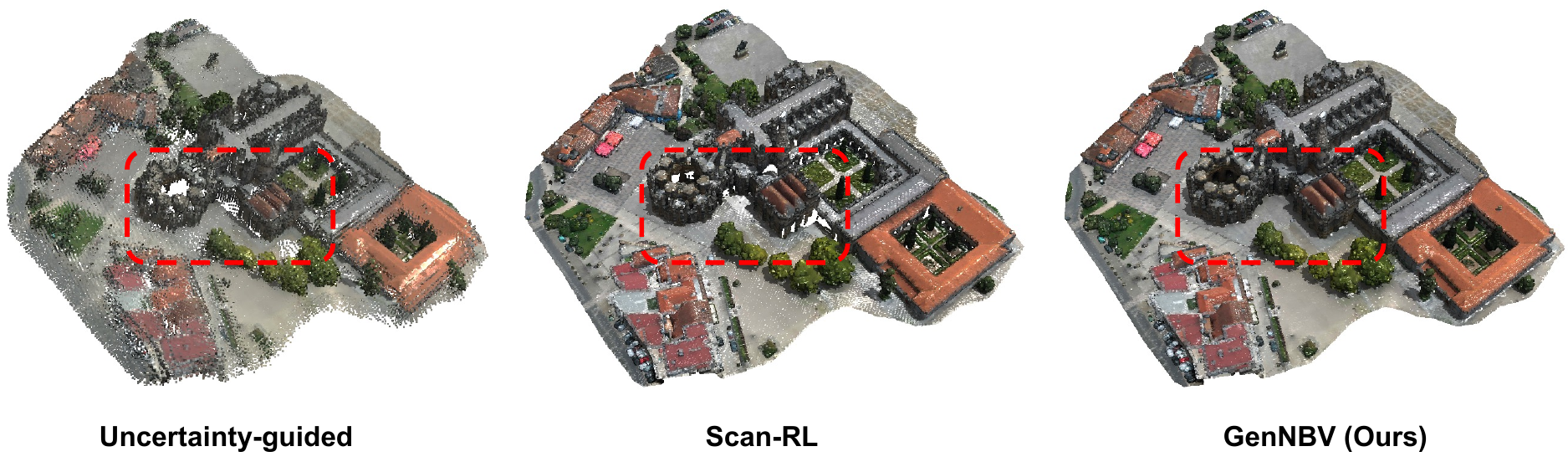}
      \caption{\small The visualization results of an unseen 3D outdoor scene with enormous details from Objaverse, reconstructed by Uncertainty-guided, Scan-RL and our model.
      Compared to the uncertainty-guided method and Scan-RL, the scene reconstructed by our method is more watertight and has fewer holes on the ground and building surface, especially in the region highlighted by the red box.}
      \label{fig:visual_obj_city}
    \end{figure*}

    As shown in Table~\ref{tab:main_table}, our GenNBV shows the best in-distribution and out-of-distribution generalizability in both coverage sufficiency and view efficiency, when evaluated on test sets consisting of unseen objects from Houses3K and OmniObject3D. In particular, the comparison in Table~\ref{tab:main_table} demonstrates the strength of the proposed representations and action space. Note that the Scan-RL's representation (`*' in the table) uses 6 frames, while our semantic representation only needs 2 frames.
    Moreover, learning-based NBV policies such as GenNBV and Scan-RL, with much larger action space, outperform all rule-based baselines, even if ActiveRMap and Uncertainty-Guided baselines are trained and evaluated on the same 3D objects.

    \noindent \textbf{Non-house Generalization} In Table~\ref{tab:non_house}, we introduce an experiment to evaluate the generalizability of our GenNBV trained on Houses3K to non-house categories. The targets include $74$ \textit{animals}, $43$ \textit{trucks} and $33$ \textit{dinosaurs} from OmniObject3D, and a challenging \textit{indoor scene (Room 0)} from Replica. The number of views is set to 20.

    Our GenNBV almost outperforms Scan-RL in all metrics. Particularly, on Replica Room 0, which consists of numerous multi-sized objects with complex occlusion, our GenNBV is significantly better than Scan-RL. This shows the effectiveness of our free-space NBV policy.

\subsection{Ablation Study}
    We implement the ablation study about multi-source state embedding and depth-based representation. Please refer to the Appendix for further experiments.
    
    \noindent\textbf{Multi-source State Embedding.}
    In Sec.~\ref{sec:method_state}, we introduce how to build our multi-source state embedding. Thanks to the same shape of representations and state embedding, we only need to adjust the input dimension of the linear layer of state embedding according to the modal type when ablating the model.
    Here we reveal the importance of specific representations with the ablation results shown in Table~\ref{tab:ablation_repre}. In uni-source experiments, we design the 5-DoF Pose baseline that learns an empirically optimal trajectory, which helps understand the effectiveness of domain knowledge of the dataset.
    We also evaluate different combinations of representation categories in Table~\ref{tab:ablation_repre}, demonstrating the effectiveness of our multi-source state embedding for policy learning in multi-source experiments.

    \vspace{-2pt}
    \begin{table}[h!]
        \huge
        \center
        \setlength{\tabcolsep}{12pt}
        \renewcommand{\arraystretch}{1.5}
        \caption{Ablation studies of depth-based representations on unseen Houses3K test set, where depth map is the only sensory source.}
        \label{tab:ablation_repre_depth}
        \resizebox{1.0\linewidth}{!}{
        \begin{tabular}{l|cc}
        \toprule
            \textbf{Depth-based Representation} & \textbf{Mean AUC} & \textbf{Final Coverage Ratio} \\ \hline
            Geometric 2D Map & 71.86\% & 74.88\% \\
            Binary 3D Grid & 77.28\% & 80.85\% \\
            \textbf{Probabilistic 3D Grid} & \textbf{81.06\%} & \textbf{84.56\%} \\
        \bottomrule
        \end{tabular}}
    \end{table}

    \vspace{1pt}
    \noindent\textbf{Depth-based Representation.}
    Furthermore, we ablate the depth-based representations in Table~\ref{tab:ablation_repre_depth}. The experimental results demonstrate that our probabilistic 3D grid is more comprehensive than a binary 3D grid and a geometric 2D map to support the NBV prediction. 




\subsection{Qualitative Results}
    We visualize the reconstruction results generated from the scanning trajectory of a single episode in Fig. \ref{fig:visual}. 
    It demonstrates that our next-best-view policy can reconstruct objects better in terms of completeness and appearance quality compared to Scan-RL.
    We further visualize the reconstruction results of an outdoor scene from Objaverse using 30 collected views in Fig. \ref{fig:visual_obj_city}. 

\section{Conclusion}
\label{sec:conclusion}

This study presents an end-to-end approach for active 3D scene reconstruction, reducing the need for manual intervention. 
Specifically, the learning-based policy explores how to reconstruct diverse objects in the training stage and thus can generalize to reconstruct unseen objects in a fully autonomous manner.
Our controller maneuvers in free space and selects the next best view based on a hybrid scene representation which conveys scene coverage status and thus reconstruction progress. 
We show the effectiveness of our approach by generalizing it on multiple datasets. 
The quantitative and qualitative generalization results on holdout Houses3K test set and cross-domain OmniObject3D, including house category and non-house categories, show that our method outperforms other baselines in terms of reconstruction completeness, efficiency and accuracy. 
Furthermore, the experiment conducted on Objaverse and Replica shows that the policy trained in house-only settings can even generalize to complicated outdoor and indoor scenes.

\vspace{4pt}
\section*{Acknowledgements}
This work is supported by Shanghai Artificial Intelligence Laboratory and CUHK Direct Grants (RCFUS) 4055189. We're very grateful to Tao Huang, Zeqi Xiao, Junfeng Long, Mulin Yu, and all reviewers for their insightful advice.

\newpage

{
    \small
    \bibliographystyle{ieeenat_fullname}

}

\clearpage
\maketitlesupplementary
\appendix

\section{Implementation Details}
\subsection{Occupancy Grid Mapping Algorithm}
\label{app:grid}
     Before updating the probabilistic occupancy grid $F^G_t$, Bresenham's line algorithm is implemented to cast the ray path in 3D space between the camera viewpoint and the endpoints among the point cloud back-projected from $D_{t+1}$. 
     According to the classical occupancy grid mapping algorithm~\citep{thrun2002probabilistic}, we have the log-odds formulation of occupancy probability:
    \begin{equation}
        \label{eqn:OGM}
        \log Odd(v_i|z_j) = \log Odd(v_i) + \log\frac{p(z_j|v_i=1)}{p(z_j|v_i=0)},
    \end{equation}
    where $v_i$ denotes the occupancy probability of $i^{th}$ voxel in the grid $F^G_t$, $z_j$ is the measurement event that $j^{th}$ camera ray passes through this voxel. 

    For the item $C=\log\frac{p(z_j|v_i=1)}{p(z_j|v_i=0)}$, there are only two cases for the measurement event in fact: $z_j=0$ or $z_j=1$. Thus, if the measurement event $z_j$ (i.g., the voxel is passed through by the $j^{th}$ camera ray) happens, we’ll update the occupancy by adding the value of $C_1=\log\frac{p(z_j = 1 | v_i = 1)}{p(z_j = 1 | v_i = 0)}$. If it’s not passed, we’ll add the value $C_2 = \log\frac{p(z_j = 0 | v_i = 1)}{p(z_j = 0 | v_i = 0)}$. The values of $C_1$ and $C_2$ can be set as an empirical constant, depending on factors such as the accuracy of ray casting and the confidence of each ray. Actually, $C’ = |\frac{C_1}{C_2}|$. We set a high incremental value for $C'$ (i.e., high confidence) because our experiments are based on the realistic simulator and accurate observations like depth maps.


    Therefore, we can update the occupancy status of each voxel in the grid $F^G_t$ by adding a constant for each ray casting process. Note that the probabilistic occupancy grid $F^G$ is continuously updated within an episode. Finally, the occupancy status of voxels can be classified into three categories: unknown, occupied, and free, by setting an empirical threshold.
    
    To demonstrate the robustness and optimality of hyper-parameter in our implementation, we show the experimental results in~\cref{tab:ablation_param}.

    \begin{table}[ht!]
        \center
        \setlength{\tabcolsep}{8pt}    
        \renewcommand{\arraystretch}{1.5}   
        \caption{Evaluation results for different empirical parameters in the implementation of occupancy grid mapping algorithm. Note that we use a constant $C'$ to represent the value of a ratio $|\frac{C_1}{C_2}|$ introduced in \ref{app:grid}}.
        \label{tab:ablation_param}
        \resizebox{1.0\linewidth}{!}{
        \begin{tabular}{c|ccc}
        \toprule
            \textbf{Occupancy Threshold} & \textbf{Mean AUC $\uparrow$} & \textbf{Final Coverage Ratio $\uparrow$} & \textbf{Accuracy $\downarrow$} \\ \hline
            \textbf{0.5 (Ours)} & \textbf{87.39\%} & \textbf{95.63\%} & \textbf{0.37} \\
            1.0 & 84.41\% & 93.43\% & 0.41 \\
            1.5 & 84.84\% & 95.20\% & 0.39 \\
            2.5 & 54.77\% & 58.70\% & 0.85 \\ \hline
            \textbf{Value of Constant $C'$} & \textbf{Mean AUC $\uparrow$} & \textbf{Final Coverage Ratio $\uparrow$} & \textbf{Accuracy $\downarrow$} \\ \hline
            5 & 66.96\% & 71.92\% & 0.49 \\
            10 & 80.75\% & 92.16\% & 0.43\\
            \textbf{20 (Ours)} & \textbf{87.39\%} & \textbf{95.63\%} & \textbf{0.37} \\
            40 & 86.97\% & 94.12\% & 0.40\\
        \bottomrule
        \end{tabular}}
    \end{table}

\vspace{5pt}
\subsection{Network and Training}
    \noindent \textbf{Training.}\quad
    In Isaac Gym~\citep{makoviychuk2021isaac}, we set 256 parallel environments for policy training, where each environment corresponds with one building-level object. 
    Once the coverage ratio reaches a threshold (99\% in practice), a collision occurs, or the episode length reaches the maximum threshold (100 steps), the environment is reset and the building to be reconstructed is replaced.
    Our NBV policy is optimized through more than 32 million iterations and uses approximately 24 hours of training time on an NVIDIA V100 GPU. All networks are randomly initialized and trained end-to-end.

    \noindent \textbf{Network.}\quad
    We propose a multi-source representation that has better generalizability. In particular, we first build two mid-level representations: a 3D geometric representation $F^G$ from depth maps and a semantic representation $F^S$ from RGB images. The geometric representation $F^G$ encoded from depth images is with the shape of $(N, X, Y, Z, 4)$, where the number of parallel training environments $N$ is 256, the grid size $(X, Y, Z)$ is $(20, 20, 20)$, and the last dimension represents the 3D world coordinate and occupancy possibility. Semantic representation $F_S$ is stacked grayscale images encoded from multi-view RGB images with the shape of $(N, M, H, W)$, where the number of RGB images $M$ is 5, the size of grayscale images is $(64, 64)$

    Specifically, we encode mid-level representations to embeddings: $s^S_t=f^S(F^S_t)$ and $s^G_t=f^G(F^G_t)$. $f^S$ encompasses a 2-layer 2D convolution and $\mathrm{Linear}(\mathrm{Flatten}(x))$ operation, while $f^G$ encompasses a 2-layer 3D convolution and $\mathrm{Linear}(\mathrm{Flatten}(x))$. Subsequently, we combine them with the historical action embedding $s^A_t=\mathrm{Linear}(a_{1:t})$ to generate the final state embedding $s_t$, as the input to the policy network. This process can be formulated as: 
        \begin{equation}
        \label{eqn:state}
        s_t = \mathrm{Linear}(s^G_t; s^S_t; s^A_t),
        \end{equation}
    where all embeddings are 256 vectors.


    Our policy network PPO, implemented by Stable Baseline3~\citep{raffin2021stable}, is a 3-layer multi-layer perceptron (MLP). The output of our policy is used to parameterize a distribution over our 5-dimensional action space. In this way, the action can be drawn from the stochastic policy $a \sim \pi(\cdot |o_t)$.

\subsection{Baseline Policies}
\label{app:baseline}
    The implementation details of baseline policies are described below:

    \noindent1) \textbf{Random Policy}: This policy randomly generates 5-dim vector $(x, y, z, pitch, yaw)$ among the action space as the next action. The randomly generated positions are constrained so as not to cause collisions. The reported results are evaluated on the test set and averaged over random seeds from 0 to 4. 

    \noindent2) \textbf{Random Policy on the Sphere}: This policy randomly generates positions $(x, y, z)$ on a pre-defined hemisphere that exactly covers all objects of the test set. The headings are required to point to the center of the hemisphere. To avoid collisions, we set the radius of the hemisphere to 9 meters, which is greater than the maximum height of the test set object. The reported results are evaluated on the test set and averaged over random seeds from 0 to 4. 
    
    \noindent3) \textbf{Uniform Policy on the Sphere}: All positions are evenly distributed on the previously mentioned hemispheres. Specifically, for Houses3K test set, all sampling points are distributed over 5 heights, each with 6 evenly spaced positions. For OmniObject3D, all sampling points are distributed over 4 heights, each with 5 evenly spaced positions.
    
    \noindent4) \textbf{Uncertainty-Guided}~\citep{lee2022uncertainty}: This NBV policy iteratively selects the next view from a pre-defined viewpoint set based on the entropy-based uncertainty based on a continually optimized neural radiance field. We use TensoRF~\citep{chen2022tensorf} as the implementation foundation of neural radiance field. Before implementing uncertainty-driven viewpoint selection, we uniformly sample 100 views on the pre-defined hemisphere as the viewpoint candidate set. Note that we need to sample the candidate set for each object.

    \noindent5) \textbf{ActiveRMap}~\citep{zhan2022activermap}: The working pipeline is similar to uncertainty-guided policy. In particular, we implement the ``discrete (free)'' setup of ActiveRMap, which constrains the drone agent on the pre-defined hemisphere. Having considered collision avoidance in the design of the viewpoint set, we remove the collision penalty in its optimization objective.

    \noindent6) \textbf{Scan-RL}~\citep{peralta2020next}: This RL-based policy predicts the next viewpoint from a 3-dim hemisphere space, relying on the historical RGB.

    \noindent7) \textbf{Ours with Scan-RL's Representation}: To further compare the former Next-Best-View policy Scan-RL~\citep{peralta2020next} with us, we implement Scan-RL in our experimental setup, with our action and state space. This free-space NBV policy uses Scan-RL's representation, which is only extracted from RGB images instead of our original hybrid representations.

\section{Data Preprocessing}
    \noindent \textbf{3D Mesh.}
    To boost our NBV policy's generalizability on building-scale objects, we rescale the original 3D meshes from Houses3K~\citep{houses3k}, OmniObject3D~\citep{wu2023omniobject3d} and Objaverse~\citep{deitke2023objaverse} to a reasonable building size. For example, the size of building meshes from Houses3K and OmniObject3D are approximately $(15m, 15m, 8m)$.

    \noindent \textbf{Ground-truth Point Cloud.}
    We use the Poisson Disk sampling method~\citep{yuksel2015sample} implemented by Open3D~\citep{Zhou2018} to sample $100,000$ points from 3D meshes. And then we voxelize these point clouds. These voxelized points are viewed as ground-truth point clouds of these meshes.

\section{Additional Experiments}
\begin{figure}[htbp]
    \vspace{-5pt}
      \centering
      \includegraphics[width=0.5\textwidth]{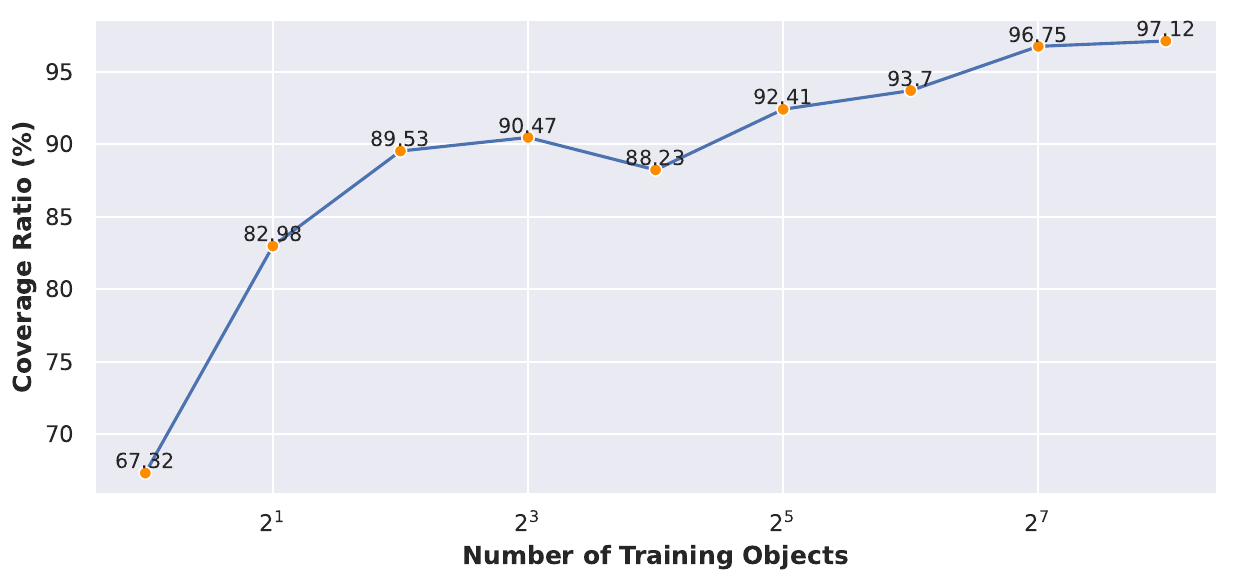}
      \caption{The curve of coverage ratio with the increasing number of training objects on unseen OmniObject3D house category.}
      \label{fig:scaling}
    \end{figure}

    \noindent\textbf{Number of Training Objects.} Motivated by generalizable RL-based policies~\citep{cobbe2020leveraging, li2023scenarionet, li2022metadrive}, we explore the impact of diversity of training data for generalizability. As shown in Table.~\ref{tab:ablation_obj} and Fig.~\ref{fig:scaling}, we found that increasing the diversity of training objects indeed leads to better generalizability for 3D reconstruction.
    \vspace{-2pt}
    \begin{table}[htbp]
    \center
    \setlength{\tabcolsep}{12pt}
    \renewcommand{\arraystretch}{1.1}
    \captionof{table}{Ablation studies of the number of training objects in our framework on unseen OmniObject3D house category.}
    \vspace{-5pt}
    \label{tab:ablation_obj}
    \resizebox{1.0\linewidth}{!}{
    \begin{tabular}{c|cc}
    \toprule
        \textbf{\makecell{Number of \\ Training Objects}} & \textbf{Mean AUC} & \textbf{Final Coverage Ratio} \\ \hline
        1 & 61.76\% & 67.32\% \\
        2 & 71.03\% & 82.98\% \\
        4 & 70.72\% & 89.53\% \\
        8 & 73.71\% & 90.47\% \\
        16 & 73.73\% & 88.23\% \\
        32 & 82.39\% & 92.41\% \\
        64 & 82.09\% & 93.70\% \\
        128 & 87.21\% & 96.75\% \\
        \textbf{256} & \textbf{88.63\%} & \textbf{97.12\%} \\
    \bottomrule
    \end{tabular}}
    \end{table}

    \vspace{-5pt}
    \noindent \textbf{The evidence for sufficient and efficient capturing.} To evaluate both completeness and efficiency, we use the area under the curve (AUC) of coverage ratio as our main metric, stated in Sec. 4.1. Additionally, the figure below shows the mean AUC curves for OmniObject3D houses. Our GenNBV achieves a better coverage ratio under 20 views (97.12\% v.s. 92.53\%), and even surpasses the saturated coverage of Scan-RL using merely 5 views.
    
    \vspace{0.2cm}
    \noindent\includegraphics[width=0.49\textwidth, height=2.5cm]{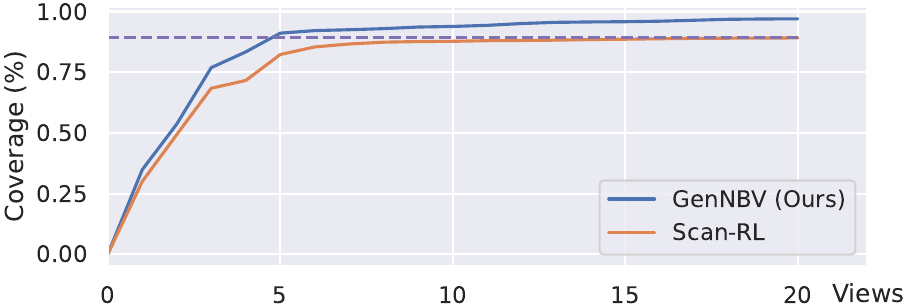}
    \vspace{-0.2cm}

    \noindent \textbf{The effectiveness of 2D semantic representations.} Below, we add the experiments on Houses3K test set to illustrate the effectiveness of the proposed semantic representation. Note that Scan-RL's representation uses 6 frames, while our semantic representation only needs 2 frames.

    \vspace{2pt}
    \noindent\resizebox{\linewidth}{!}{%
        \centering
        \begin{tabular}{c|l|ccc}
        \hline
        \textbf{Num. of RGB} & \textbf{NBV Policy} & \textbf{AUC $\uparrow$} & \textbf{FCR $\uparrow$} & \textbf{Acc. $\downarrow$}\\
        \toprule
        \multirow{2}{*}{2} & Ours w/ Scan-RL Repre. & 76.31\% & 79.35\% & 0.50\\
         & Ours (RGB-only) & \textbf{81.24\%} & \textbf{87.90\%} & \textbf{0.45} \\ \hline
        \multirow{2}{*}{6} & Ours w/ Scan-RL Repre. & 87.39\% & 95.63\% & \textbf{0.38} \\ 
         & Ours (RGB-only) & \textbf{87.95\%} & \textbf{96.92\%} &\textbf{0.38} \\ \hline
        \end{tabular}%
    }\vspace{0.1cm}

    In addition, we preprocess the RGB images to grayscale images because we empirically find that the grayscale images featuring edges are sufficient to guide the NBV prediction and achieve slightly better performance.

\end{document}